\documentclass[letterpaper]{article} % DO NOT CHANGE THIS
\usepackage{aaai23}  % DO NOT CHANGE THIS
\usepackage{times}  % DO NOT CHANGE THIS
\usepackage{helvet}  % DO NOT CHANGE THIS
\usepackage{courier}  % DO NOT CHANGE THIS
\usepackage[hyphens]{url}  % DO NOT CHANGE THIS
\usepackage{graphicx} % DO NOT CHANGE THIS
\usepackage{natbib}  % DO NOT CHANGE THIS AND DO NOT ADD ANY OPTIONS TO IT
\usepackage{caption} % DO NOT CHANGE THIS AND DO NOT ADD ANY OPTIONS TO IT
\usepackage{algorithm}
\usepackage{algorithmic}
\usepackage{adjustbox}
\usepackage{multirow}
\usepackage{multicol}
\usepackage{booktabs}       % professional-quality tables
\usepackage{amsfonts}
\usepackage{bbm}
\usepackage{amsmath}
\usepackage{subfigure}
\usepackage{hyperref}
\usepackage[dvipsnames]{xcolor}
\usepackage{arydshln}

\DeclareMathOperator*{\minimize}{minimize}

\usepackage{newfloat}
\usepackage{listings}
\DeclareCaptionStyle{ruled}{labelfont=normalfont,labelsep=colon,strut=off} % DO NOT CHANGE THIS
\floatstyle{ruled}
\newfloat{listing}{tb}{lst}{}
\floatname{listing}{Listing}
\title{Everyone's Voice Matters: Quantifying Annotation Disagreement \\Using Demographic Information}
\author{
    Ruyuan Wan\textsuperscript{\rm 1}\footnote{This work was done while RW and JK were at the Minnesota NLP lab.}, \hspace{2mm}
    Jaehyung Kim\textsuperscript{\rm 2}$^*$,
    \hspace{2mm}
    Dongyeop Kang\textsuperscript{\rm 3}
}
\affiliations{
    \textsuperscript{\rm 1}University of Notre Dame, \textsuperscript{\rm 2}KAIST,
    \textsuperscript{\rm 3}University of Minnesota
    \\
    rwan@nd.edu, jaehyungkim@kaist.ac.kr, dongyeop@umn.edu
}

\usepackage{bibentry}
\usepackage{multirow}
\usepackage{makecell}

\begin{document}

\maketitle

\begin{abstract}
In NLP annotation, it is common to have multiple annotators label the text and then obtain the ground truth labels based on the agreement of major annotators. 
However, annotators are individuals with different backgrounds, and minors' opinions should not be simply ignored.
As annotation tasks become subjective and topics are controversial in modern NLP tasks, we need NLP systems that can represent people's diverse voices on subjective matters and predict the level of diversity.
This paper examines whether the text of the task and annotators' demographic background information can be used to estimate the level of disagreement among annotators.
Particularly, we extract disagreement labels from the annotators' voting histories in the five subjective datasets, and then fine-tune language models to predict annotators' disagreement. 
Our results show that knowing annotators' demographic information, like gender, ethnicity, and education level, helps predict disagreements. 
In order to distinguish the disagreement from the inherent controversy from text content and the disagreement in the annotators' different perspectives, we simulate everyone's voices with different combinations of annotators' artificial demographics and examine its variance of the fine-tuned disagreement predictor. 
Our paper aims to improve the annotation process for more efficient and inclusive NLP systems through a novel disagreement prediction mechanism. 
Our code and dataset are publicly available.
\footnote{\url{https://github.com/minnesotanlp/Quantifying-Annotation-Disagreement}}
\end{abstract}

\section{Introduction}

%present the topic: subjective annotation disagreement, two types of controversy (text, annotators), demographics
Supervised AI systems are trained on annotated datasets with labels determined by consensus among multiple annotators.
The subjective opinions of different annotators often bring annotation disagreement in the decision of the final labels.
% Subjective annotation is usually made by crowd workers, which sometimes brings annotation disagreement in final labels.
% To quantify the subjective study, researchers usually take aggregated labels from annotators to achieve a gold label for training supervised machine learning models \cite{davani2022dealing}. 
Most commonly, this disagreement is addressed by ignoring highly-disagreed cases and only including those whose opinions were voted on by the majority as the final label.
% However, this may ignore the disagreement among the annotators. 
When the labeling tasks become more subjective and require the annotator's own interpretation and judgment, such as detecting offensiveness and judging social dilemmas \cite{Reidsma2008ExploitingA}, this majority-based aggregation often fails to learn the true distribution of annotators' voices.
% Furthermore, labeling subjective tasks, such as sentiment, offensiveness, and social dilemmas, requires the annotator's interpretation and judgment \cite{Reidsma2008ExploitingA}.  
The increasing subjectivity of NLP problems in modern NLP will cause the annotators' disagreement not only due to the potential random error in the process but also because annotators may interpret the text with different views and make judgments based on their own connotations. 

Different demographics, cultural backgrounds, and living experiences influence how people receive and interpret information. This difference is more visible in subjective tasks. For example, Sap et al. found more consecutive annotators who have higher scores on racist beliefs are more likely to label African American English as toxic rather than label anti-Black language as toxic \cite{ sap-etal-2022-annotators}. In these cases, the aggregated singular labels can bring bias by using less inclusive and \textit{societal-representative labels} to accommodate everyone's voices in subjective studies.

\begin{figure}[t]
\centering
\includegraphics[width=0.49\textwidth,trim={0.7cm 0.5cm 0.7cm 0.7cm},clip]{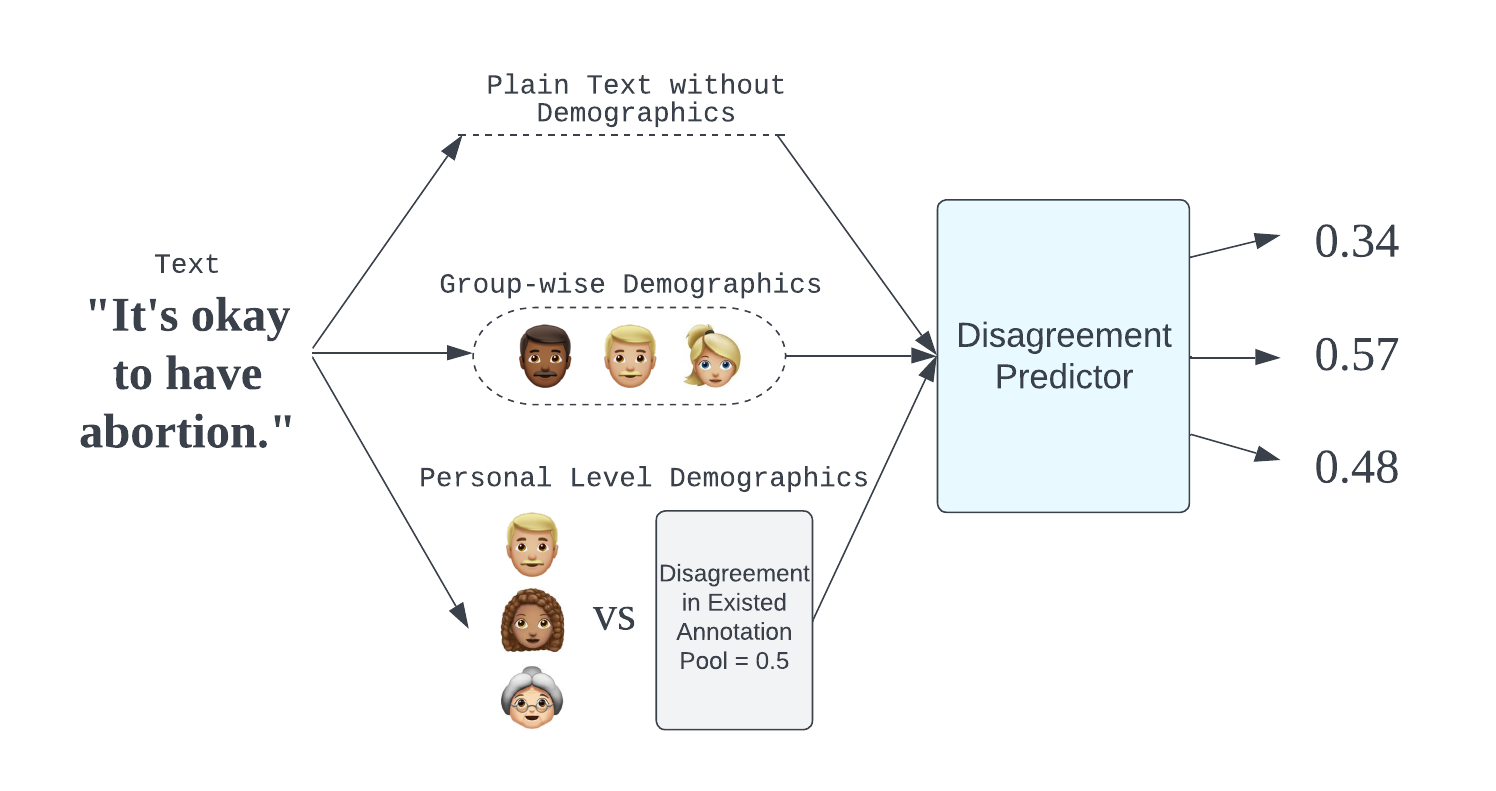}
\caption{Disagreement prediction predicts disagreement only from input text or input text with (group-wise or individual against the majority) annotators' demographic information.}
\label{fig:disagreement_concept}
\end{figure}

%describe the background: misleading bias in using low-quality final labels to represent all people
% Motivated by \cite{ sap-etal-2022-annotators}, w
This paper assumes that annotators' disagreement potentially comes from the \textit{limited representations of the annotator group assigned} or \textit{controversy of the text in nature}. 
This study focuses on exploring the relationship between the annotator group and natural controversy in text by developing a \textbf{disagreement predictor} with and without the information about the annotator group, as depicted in Figure \ref{fig:disagreement_concept}. 
% establish the research problem:
In particular, we analyze annotators' disagreement from five subjective task datasets to answer the following research questions: 
\begin{itemize}
\item[Q1:]
Is it possible to predict the level of annotators' disagreement with text using language models? Does knowing annotators' identities, like demographic information in addition to text, help predict annotation disagreement? 
\item[Q2:] 
Is the disagreement caused by the natural controversy of the text or by the biased distribution of the assigned annotators? 
\end{itemize}

% Map out the paper
% 1. disagreement prediction: Hard(binary) v.s. Soft(continuous)
% 2. Demographics Augmentation: 
%       Group identities v.s. Personal identity
%       Sentence format v.s. Column format
% 3. Simulation demographics 
% 4. Human Evaluation
Our research demonstrates that disagreement is  predictable in subjective annotation tasks by using Roberta model \cite{liu2019roberta} to predict both hard disagreement (binary label) and soft disagreement (continuous label). We further design two demographic augmentation experiments and find that bringing the annotator-level demographic information can significantly improve disagreement prediction performance. Finally, based on the findings, we simulated artificial annotators' backgrounds to predict disagreement to check whether the disagreement score will be changed in a wider annotator population.
%; e.g., if the variance between all simulated annotators is high, the corresponding text's disagreement is dependent on the limited representation of the group of annotators. 
%Conversely, the disagreement is more related to the controversy of content. %add human evaluation here if we can accomplish it 
% In short, we propose a disagreement measurement that can optimize annotators' efforts by estimating the disagreement source and identifying the difficult cases that require more attention, also helping improve the fairness and quality of subjective annotation.
In short, we propose a disagreement measurement that can efficiently suggest the optimal number of annotators and assign an appropriate demographic group of annotators per text, possibly helping improve the fairness and quality of subjective annotation.
% optimize annotators' efforts by estimating the disagreement source and identifying the difficult cases that require more attention, also helping improve the fairness and quality of subjective annotation.

% specify the objectives: identify the annotation disagreement to improve annotation quality

\section{Related Works}
%subjective tasks
%(dis)agreement labels
%demographics

%knowing what the challenges are 
Tasks like toxicity detection \cite{Sap2020SocialBF, Yu2022HateSA}, sentiment analysis \cite{Potts2021DynaSentAD}, and social, ethical labeling \cite{Forbes2020SocialC1, Hendrycks2021AligningAW} are highly subjective and controversial. One can think one post is offensive, but others may consider it acceptable. There is no one objective ground truth. \citeauthor{Rttger2022TwoCD} summarized three key challenges of descriptive annotation in subjective NLP tasks: interpretation of disagreement, label aggregation, and representativeness of annotators.

% interpret disagreement
Due to the absence of absolute ground truth, the interpretation of disagreement becomes complicated \cite{Alm2011SubjectiveNL}. For instance, the disagreement may result in considerably different reliabilities: whether the annotators disagree on the most critical or least crucial instances \cite{Foley2018ExplainableAT}. In addition, researchers commonly use some notion of the agreement to measure the task's subjectiveness, such as using inter-annotator agreement metrics Cohen's Kappa \cite{cohen1960coefficient} or Fleiss' Kappa \cite{fleiss1971measuring} to measure annotations' reliability. But when presenting the final results in the downstream task, people usually use the aggregated labels that can conceal informative disagreement and evaluation metrics that are unaware of the task's subjectiveness \cite{Rttger2022TwoCD}. 

% label aggregation and possible solutions 
Rather than 'correct' or 'wrong,' \citeauthor{Alm2011SubjectiveNL} pointed out the concept of acceptability. There might be multiple acceptable answers in subjective tasks. However, aggregating labels through major voting will increase the risk of discarding minority voices. To address the problem, \citeauthor{davani2022dealing}  treats predicting each annotator's judgments as separate subtasks, which achieved the same or better performance than aggregating labels in the data before training. They also further evaluate the model uncertainty using the variance of the predicted annotation label. However, this still concerns that recognizing the aggregated major votes as the final targets does not always represent all acceptable answers. On the other hand, \citeauthor{Uma2022ScalingAD} used posterior calibration with a soft-loss approach to learning from data containing disagreement. They noticed that temperature scaling only functions with data where disagreements are caused by label overlap and not with data where disagreements are caused by annotator subjective judgment or language ambiguity. This aligns with \citeauthor{Foley2018ExplainableAT} 's finding for tasks with subjective labels: without collecting additional labels, models reach the ceiling of performance given the small dataset size and the inherent disagreement between annotators on which documents are controversial.

% demographics: representativeness of annotators
In previous research, annotators' demographics have shown importance in improving the annotation quality in subjective tasks. For example, \citeauthor{Prabhakaran2021OnRA} demonstrated that the agreement scores could be very significant among different socio-demographic groups annotators identified when certain individual annotators disagree with the majority labels. Further, \citeauthor{Gordon2022JuryLI} proposed jury learning, a recommender system approach defining which people or groups, in what proportion, determine the classifier's prediction. For instance, a jury learning model would recommend women and black jurors for online hate speech detection, who are mainly targeted in online harassment.  However, many public datasets didn't collect annotators' demographics with their annotations. Further, the datasets that reported the annotator's demographics also have imbalanced representative concerns. For example, the race is often skewed, and dominant with the white race \cite{Sap2020SocialBF, Forbes2020SocialC1, Hendrycks2021AligningAW, sap-etal-2022-annotators}.  

As shown above, researchers have implicitly resolved the label disagreement using majority votes, annotator selection, and the soft-loss approach\cite{Uma2021SemEval2021T1}. Different from the interrater disagreement resolution, which defines disagreement as a sign of poor quality or mistakes to be resolved\cite{oortwijn2021interrater}, our research explicitly quantifies disagreement as our task target and further distinguishes the nuance among various socio-demographic groups. 
%Emphasize the Significance of the problem: The social impact problem considered by this paper is significant and has not been adequately addressed by the AI community. 

\section{Methods}\label{sec:method}

% Overview
This section presents our method for quantifying subjective annotation disagreements. 
% One more sentence? : intuition behind this
Our main idea is modeling the annotation disagreement using demographic information of each annotator as additional inputs, with the pre-trained language model, \textit{e.g.,} RoBERTa \cite{liu2019roberta}. 
In Section \ref{sec:3.1}, we first introduce the mathematical notations. Then, we elaborate on the details of the proposed method in Section \ref{sec:3.2}. Finally, we provide a way to simulate the annotators' demographic information in Section \ref{sec:3.3}.

\subsection{Preliminaries}
\label{sec:3.1}

\begin{figure*}[h]
\centering
\includegraphics[width=0.8\textwidth,trim={0 0.7cm 0 0},clip]{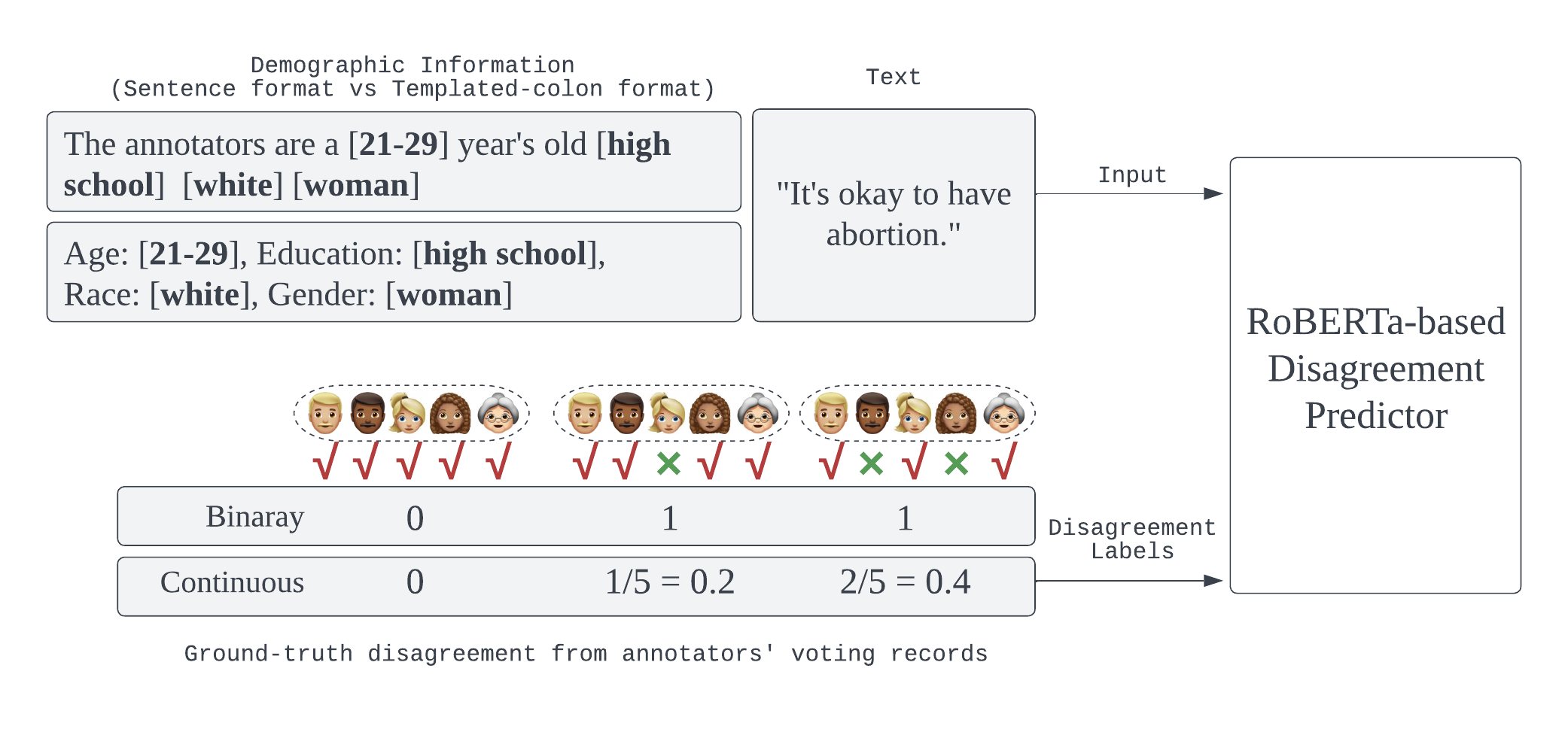}
\caption{Our proposed disagreement predictor that takes the task sentence and/or (group or person) demographic information as input and ground-truth disagreement among annotators as labels. The demographic information is concatenated to the task sentence either in sentence format or templated-colon format. The ground-truth labels are aggregated from the annotators' voting records as binary labels with a threshold (i.e., 3/5) or continuous labels as they are.}
\label{fig:disagreement_formulation}
\end{figure*}

We first describe the problem setup of our interest under a text classification scenario with $K$ classes.
Let $\mathcal{D}=(\mathcal{X}, \mathcal{Y})$ denote the given annotated dataset where $\mathcal{X}$ is a set of texts and $\mathcal{Y}$ is the annotation matrix of $\mathcal{X}$. 
Specifically, each entry of $\mathcal{Y}$, $y_{i}(\mathbf{x}) \in \{1, \dots, K\}$, represents $i$th annotation assigned to text $\mathbf{x} \in \mathcal{X}$.\footnote{We clarify that the $i$th annotation could be labeled by different annotators between different texts in $\mathcal{X}$.}
We assume that there are $N$ different annotations for each text $\mathbf{x}$ and $\mathbf{y}(\mathbf{x}) = [y_{1}(\mathbf{x}), \dots, y_{N}(\mathbf{x})]$ denotes all annotations assigned to $\mathbf{x}$. 
Then, $r_{k}(\mathbf{x})=\sum_{i=1}^{N} \mathbbm{1}[y_{i}(\mathbf{x}) = k] / N$ denotes the agreement rate of $\mathbf{x}$ to the label $k$ where $\sum_{k=1}^{K} r_{k}(\mathbf{x}) = 1$.
In addition, we assume that $T$ different demographic information of all $N$ annotators is available such as gender, age and race\footnote{This assumption will be relaxed in Section \ref{sec:4.5}}, and denote it as $\mathbf{d}^{\tt (t)}(\mathbf{x}) = [d^{\tt (t)}_{1}(\mathbf{x}), \dots, d^{\tt (t)}_{N}(\mathbf{x})]$ where $t=1,\dots,T$. 
Remarkably, majority voting, which is a popular common practice of assigning the label from the multiple annotations $\mathbf{y}(\mathbf{x})$ to the maximally agreed label, can be represented as $y_{\tt maj}(\mathbf{x}) := \arg \max_{k} r_{k}(\mathbf{x})$.

\textbf{Binary vs Continuous disagreement labels}.
From the agreement rate $r_{k}(\mathbf{x})$, we first compute a binary disagreement label $\bar{r}_{\tt b}(\mathbf{x}) = \mathbbm{1}[r_{y_{\tt maj}}(\mathbf{x}) \ne 1]$, which indicates if there are different opinions among the annotators for this instance $\mathbf{x}$. 
We further define a continuous disagreement label $\bar{r}_{\tt c}(\mathbf{x}) = 1 - r_{y_{\tt maj}}(\mathbf{x})$, that has the scale of 0 (everyone agrees with the same annotation result) to 1 (a significant number of people holding different opinions on the annotation results).  
Namely, the binary label $\bar{r}_{\tt b}$ indicates the existence of at least some different opinions, and the continuous label $\bar{r}_{\tt c}$ measures the degree of disagreement among the annotators. 
Without loss of generality, we refer both types of disagreement as $\bar{r}$.
The text with highest disagreement means annotators hold different opinions, and this text content is very controversial. 

\subsection{Disagreement Prediction with Demographic Information}
\label{sec:3.2}

Our goal is to predict the disagreement $\bar{r}(\mathbf{x})$ of given text $\mathbf{x}$ because it provides an effective way to understand which content is controversial or not.
To this end, our first idea is to utilize the pre-trained language model, e.g., RoBERTa \cite{liu2019roberta}, for training a predictor $f_{\theta}$ of the disagreement of given text.
Specifically, we train the model by minimizing a mean square error (MSE) loss as follow:

\begin{equation}
    \displaystyle{\minimize_{\theta} \mathcal{L}_{\tt MSE}(f_{\theta}({\mathbf{x}}), \bar{r}(\mathbf{x}))}
\end{equation}\label{eq:vanilla}

However, the annotators' disagreement is not only from the controversy of the text in nature but also from the limited representations of the assigned annotator group. Hence, more than just using text as input is needed to capture the disagreement fully.

\textbf{Incorporation of demographics: Group vs Personal}.
To this end, our key idea is incorporating the demographic information of annotators $\{\mathbf{d}^{\tt (t)}(\mathbf{x})\}_{t=1}^{T}$ to train the model $f_{\theta}$. 
Intuitively, it is expected to encode the valuable information of the disagreement of the text $\mathbf{x}$, especially related to limited representations of the annotator group assigned.
To be specific, we propose two different ways to incorporate the demographic information: (1) \textit{Text with group demographic information} and (2) \textit{Text with personal demographic information}.

\begin{table*}[t]
    \caption{Examples from the five disagreement datasets used in this paper. A stands for annotator.}
    \vspace{-4mm}
	\begin{center}
	\begin{adjustbox}{width=1\linewidth}
	\begin{tabular}{@{}c@{}c@{}c@{}c@{}}
        \toprule
        \textbf{Datasets} &  \textbf{Text} & \textbf{Annotation Distribution} & \textbf{Disagreement Label}\\
        \midrule
		\texttt{SBIC} 
		& \makecell{ ``Abortion destruction of the nuclear family  \\ contraceptives feminism convincing women to wait for   \\ children damaging economy so youth cannot leave \\ the nest ramping up tensions between sexes all \\ serves one primary goal to lower the population."} 
        & \makecell[r]{ A1 (age: 32, politics: liberal, race: white, gender: woman) \\votes for \underline{inoffensive}\\A2 (age: 34, politics: liberal, race: white, gender: woman) \\votes for \underline{inoffensive}\\A3 (age: 29, politics: mod-liberal, race: hispanic, gender: woman) \\votes for \underline{offensive} \\
        $\longrightarrow$ Aggregated Label: \textbf{inoffensive}}
         & \makecell{Binary: 1\\ Continuous: 1/3}\\ 
        \midrule
        \texttt{SChem101}  
        & \makecell{ ``It's okay to have abortion."} 
        & \makecell[r]{ A1 (age: 30-39, education: high school, race: white, gender: woman) \\votes for \underline{people ocassional think this}\\A2 (age: 40-49, education: grad, race: white, gender: man) \\votes for \underline{controversial}\\A3 (age: 30-39, education: bachelor, race: white, gender: man) \\votes for \underline{common belief}\\ A4 (age: 21-29, education: high school, race: white, gender: woman) \\votes for \underline{controversial}\\ A5 (age: 30-39	, education: bachelor, race: hispanic, gender: woman) \\votes for \underline{controversial} \\ $\longrightarrow$ Aggregated Label: \textbf{controversial}}
        & \makecell{Binary: 1\\ Continuous: 2/5 }\\
        \midrule
        \texttt{Dilemmas}  
        & \makecell{ 1st action: ``refusing to do a survey on the credit card  \\reader while paying with cash at the Office Max." \\ 2nd action: ``saying my bf has no right to dictate \\ who I tell about my abortion."} 
        & \makecell[r]{  1 annotator votes for the \underline{first action} is less ethical \\while 4 others vote the \underline{second action} is less ethical \\ $\longrightarrow$ Aggregated Label: \textbf{2nd action is less ethical}}
        & \makecell{Binary: 1\\ Continuous: 1/5 }\\
        \midrule
        \texttt{Dynasent}  
        & \makecell{ ``Had to remind him to toast the sandwich." } 
        & \makecell[r]{ 4 annotators believe it's \underline{negative} while one think it is \underline{neutral} \\ $\longrightarrow$ Aggregated Label: \textbf{negative}}
        & \makecell{Binary: 1\\ Continuous: 1/5 }\\
        \midrule
        \texttt{Politeness}  
        & \makecell{ ``Where did you learn English? \\How come you're taking on a third language?" } 
        & \makecell[r]{ 5 annotators politeness scores are \underline{5, 13, 9, 11, 11} \\ with the maximum of 25. \\ $\longrightarrow$ Aggregated Label: \textbf{impolite}}
        & \makecell{Binary: 0\\ Continuous: 0} \\
        \bottomrule
	\end{tabular}
    \end{adjustbox}
    \end{center}
    \label{table:datasets}
\end{table*}

Text with group demographic information $\widetilde{\mathbf{x}}_{\tt group}$ is constructed by listing all $N$ annotators' information $\mathbf{d}^{\tt (t)}(\mathbf{x})$ in one string and then concatenating with the targeted text $\mathbf{x}$: 
\begin{equation}
    \widetilde{\mathbf{x}}_{\tt group}=\text{Combine}[\mathbf{d}^{(1)}(\mathbf{x}), \dots, \mathbf{d}^{(T)}(\mathbf{x}), \mathbf{x}]
\end{equation}
Therefore, the group demographics supplemented text also has the same number of instances as the original dataset.

On the other hand, text with personal demographic information $\widetilde{\mathbf{x}}_{\tt person}$ is constructed by concatenating only one annotator's demographic with text:
\begin{equation}
    \widetilde{\mathbf{x}}_{\tt person}=\text{Combine}[d^{(1)}_{j}(\mathbf{x}), \dots, d^{(T)}_{j}(\mathbf{x}), \mathbf{x}]
\end{equation}
where $j=1,\dots,N$ and hence it results in $N$ times larger dataset with $N$ different annotators.

\textbf{Format: Templated vs Sentence}.
For combining the demographic information and text, we further propose two different ways with specific templates: (1) \textit{Templated format} and (2) \textit{Sentence format}.
Templated format represents the category and value of each demographic information in a separate sentence, then concatenate all of them with the given text. 
For example, if one annotator is 36 years white woman, this demographic information is converted to \textit{"Age: 36, Color: white, Gender: women"}, then concatenated with the original sentence in case of the text with person demographic.
On the other hand, sentence format represents the demographic information with a natural sentence, \textit{e.g., the annotator is a 36 years old white woman.}, then concatenate it with the original sentence.

With these demographics supplemented text $\widetilde{\mathbf{x}}$ $(\widetilde{\mathbf{x}}_{\tt group}$ or $\widetilde{\mathbf{x}}_{\tt person})$, we train our model similar to the case with the original sentence $\mathbf{x}$ in Equation (1):
\begin{equation}
    \displaystyle{\minimize_{\theta} \mathcal{L}_{\tt MSE}(f_{\theta}(\widetilde{\mathbf{x}}), \bar{r}(\mathbf{x}))}
\end{equation}

An illustration of the proposed demographic-based disagreement predictor is presented in Figure \ref{fig:disagreement_formulation}.

\subsection{Simulation of Demographic Information}\label{sec:3.3}
In addition, we propose a simulation of demographic information, which is a novel approach to analyze \textit{how the different annotator groups impact disagreement prediction}. 
It is expected to separately reveal the inherent disagreement of annotators from the controversy of the text in nature. 
Specifically, instead of ground-truth $\{\mathbf{d}^{(t)}(\mathbf{x})\}_{t=1}^{T}$, we combine the artificial demographic information $\{\bar{\mathbf{d}}^{(t)}(\mathbf{x})\}_{t=1}^{T}$ with the given text $\mathbf{x}$ and annotations $\mathbf{y}(\mathbf{x})$, to simulate the scenario with different annotators. 
Such as, the gender demographic type has \textit{four} possible options: woman, man, transgender, non-binary; and ethnicity with \textit{seven} options: white, black or African American, American Indian or Alaska Native, Asian, Native Hawaiian or other pacific islanders, Hispanic, or some other race. 
Overall, we have a total $28 = 4 \times 7$ different combinations of the annotator's demographic information for the simulation, while the ground-truth demographic information is one of them; hence, it offers an opportunity to explore the more extensive range of demographic information with the increased number of instances. 
Then, we obtain a predicted disagreement using $f_{\theta}$, which is trained with $\mathbf{x}$ and $\{\mathbf{d}^{(t)}(\mathbf{x})\}_{t=1}^{T}$ as introduced in Section \ref{sec:3.2}.

Then, we evaluate whether the predicted disagreement is easily or hard to be changed among the simulated demographic profiles so that we can distinguish whether the disagreement comes from the controversy of text or uncertainty from annotators for the disagreement label. 
For example, if the variation of predicted disagreement among the simulated combinations is high and the average change of the predicted disagreement between the simulated combinations and real disagreement is large, it might reveal that disagreement is highly related to the uncertainty of annotators. 
In contrast, the lower variation and smaller change between real disagreements indicate the disagreement is based on the controversy in the text, which is stable disagreement among various kinds of people. 

\section{Experiments}\label{sec:4}

\subsection{Benchmark Datasets}\label{sec:4.1}
To obtain the annotators' disagreement, we choose the following five datasets of subjective tasks that include annotators' voting records in the raw format.\footnote{Note that only the SBIC and SChem101 datasets report annotators' demographic information, so we used these two datasets to evaluate the effect of including demographic information in disagreement prediction.
}
% We choose public datasets of subjective tasks that contain annotators' voting records from their original raw dataset, so we can analyze annotators' disagreement. We use the following five datasets to train our disagreement predictors. 
%We choose datasets that include subjective and social tasks. We especially select the public datasets that contain annotators' voting records from their original raw dataset so that we can analyze annotators' disagreement. We use the following five datasets to train our disagreement predictors. Since only the Social Bias Inference Corpus (SBIC) and Social Chemistry 101 corpus (Schem101) report annotators' demographic information, we only used these two datasets to evaluate the effect of including demographic information in disagreement prediction. 

\textbf{Social Bias Inference Corpus (SBIC)} \cite{Sap2020SocialBF}
contains 150k structured annotations of social media posts. 
%The Social Bias Inference Corpus (SBIC) dataset contains 150k structured annotations of social media posts. 
Each post has three different annotators. Annotators indicated
%\citeauthor{Sap2020SocialBF} asked annotators to indicate 
whether the post could be considered ``offensive to anyone.'' The offensiveness is a categorical variable with three possible answers (yes, maybe, no). 

% Each social media post is annotated by three annotators.

%Our analysis finds that their annotators' pool was relatively gender-balanced and age-balanced (55\% women, 42\% men, <1\% non-binary; 36±10 years old), but racially skewed (82\% White, 4\% Asian, 4\% Hispanic, 4\% Black). And it was also politically skewed (63\% liberal, 20\% conservative). Overall, workers agreed on a post being offensive at a rate of 76\%. Later, \citeauthor{Sap2020SocialBF} showed that annotator identity and beliefs are highly related to their toxicity ratings in their annotators with attitudes paper \cite{sap-etal-2022-annotators}. 

\textbf{Social Chemistry 101 (SChem101)} \cite{Forbes2020SocialC1}
%Social Chemistry 
%described in natural language. 
is a corpus of cultural norms via free-text rules-of-thumb created by crowd workers.
%apply a bottom-up approach by asking crowd workers' cultural norms via free-text rules-of-thumb as the basic conceptual units. 
A rule-of-thumb is a judgment of action which
%, such as `It is okay to have an abortion.' Each rule of thumb 
is further broken down into 12 theoretically-motivated dimensions of people’s judgments. Our study focuses on the anticipated agreement category. It reflects workers' opinion on what portion of people probably agree with the judgment given the action. The category has five possible answers: almost no one believes, people occasionally think this, controversial, common belief, universally true. Each rule of thumb is annotated by five workers. 

%Similar to the demographic distribution in the SBIC dataset, the crowd worker pool in SChem101 is also gender-balanced and race-skewed: 55\% were women and 45\% men. 89\% of workers identified as white, 7\% as Black. 39\% were in the 30-39 age range, 27\% in the 21-29, and 19\% in the 40-49 age range. Regarding education, 44\% had a bachelor’s degree, and 36\% had some college experience or an associate's degree. However, even though some people consider one rule as a common belief, it is possible that other people think no one believes it.

\textbf{Scruples-dilemmas} \cite{Lourie2021ScruplesAC}  
%The Dilemmas dataset 
is a resource for normative ranking actions. Each instance pairs two unrelated actions and identifies which action crowd workers found less ethical. %Such as ``refusing to do a survey on the credit card reader while paying with cash at the Office Max.'' and ``saying my bf has no right to dictate who I tell about my abortion.''
Each instance is annotated by five different annotators. 

\textbf{Dyna-Sentiment} \cite{Potts2021DynaSentAD} is an English language benchmark task for ternary sentiment analysis. Each Yelp review is validated by five crowd workers into three possible sentiment results: positive, negative, and neutral. 

\textbf{Wikipedia Politeness} \cite{DanescuNiculescuMizil2013ACA}
%This dataset 
is a collection of requests from Wikipedia Talk pages, annotated with politeness. Each Wikipedia request is annotated by five annotators on a 1 to 25 scale. As \citeauthor{DanescuNiculescuMizil2013ACA} ignored neutral cases for politeness prediction, we extracted the disagreement between the binary classes of request, i.e., polite and impolite. 

\textbf{Disagreement Label Distributions}
\begin{figure}[t]
\centering
\includegraphics[width=0.45\textwidth,trim={0.9cm 0.2cm 1cm 1cm},clip]{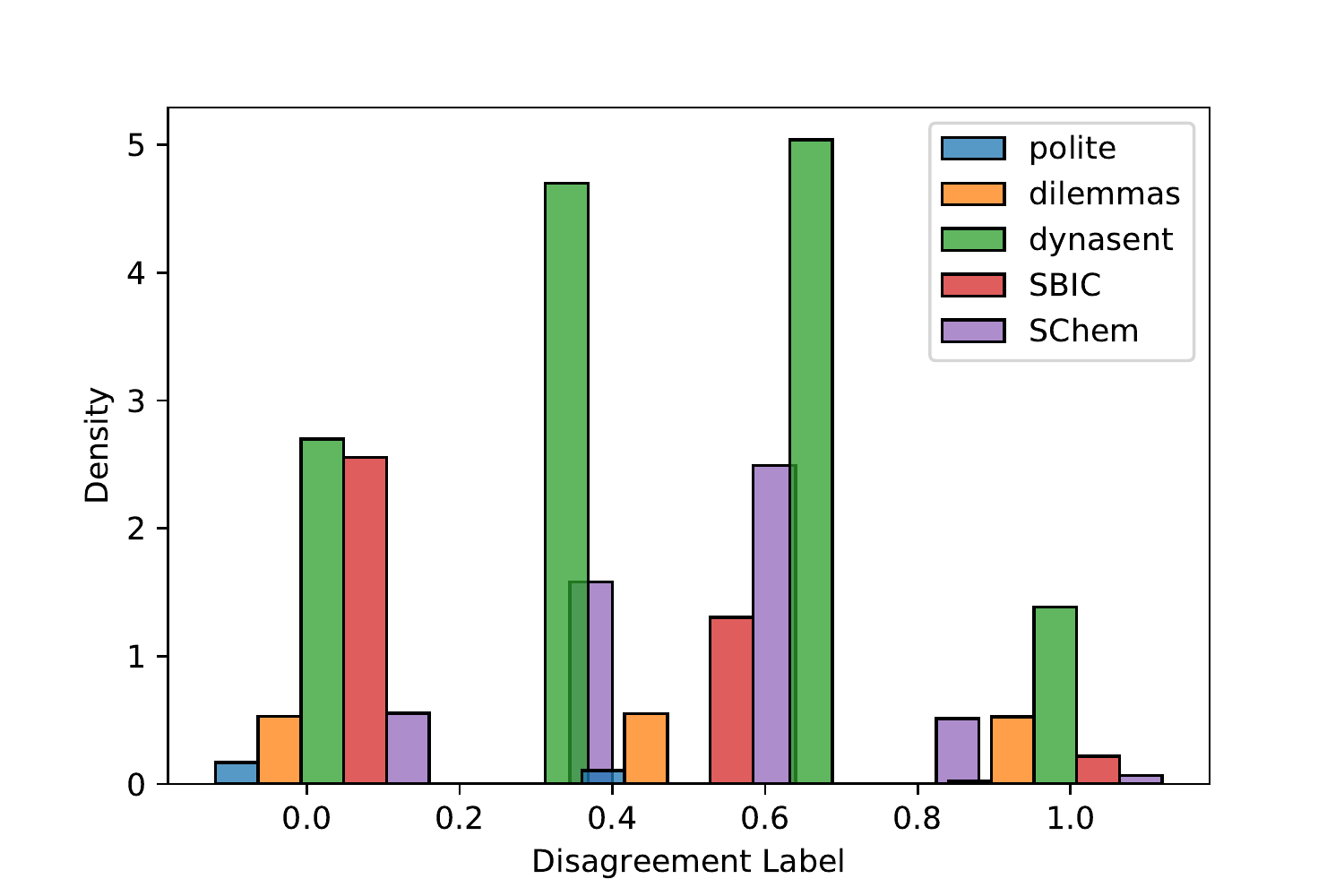}
\caption{Disagreement distributions for five datasets}
\label{fig:disagreement_distribution}
\end{figure}
The Figure \ref{fig:disagreement_distribution} shows the distributions of disagreement scores among five datasets. For dynasent dataset, since the majority of the dataset has disagreement between 0.3 to 0.6. The prediction concentrate around 0.4 to 0.5. The comparison among multiple datasets reflects that the subject topics influence the crowd annotators' disagreement. For example, most texts regarding offensiveness had consensus opinions from the annotators, while most annotators disagreed regarding sentiment.

% \iffalse
% \subsection{Disagreement Formulation}

% First, we computed a binary disagreement label for each instance in the five datasets. The binary disagreement label indicates if there are different opinions among the annotators for this instance. However, there is a range between full agreement and complete disagreement. We further define a continuous disagreement rate on the scale of 0 (everyone agrees with the same annotation result) to 1 (a significant number of people holding different opinions on the annotation results).  
  
% That is, the binary label indicates the existence of at least some different opinions, and the continuous rate measures the degree of disagreement among the annotators. 
% The most controversial case is that many people hold different opinions. The highest disagreement rate means annotators have different opinions, and the text content is very controversial. 

% we use a hard score F1 and a soft score MSE to evaluate the model performance and compare the measurement effect of binary disagreement label and continuous disagreement rate. Also, F1 and MSE evaluate the performance of language models in disagreement prediction.  
% \fi
\begin{table}[!t]
    \caption{Evaluation results of vanilla (RoBERTa) classifiers \textbf{only with text input} on the five datasets with disagreement.}
    \vspace{-4mm}
	\begin{center}
	% \begin{adjustbox}{width=0.7\linewidth}
	\begin{tabular}{@{}c cc @{\hskip 5mm} cc@{}}	
        \multicolumn{1}{c}{} &  \multicolumn{2}{c}{\textbf{Binary Label}} & \multicolumn{2}{@{}c}{\textbf{Continuous Label}}\\
        \toprule
		\textbf{Datasets}  & F1 ($\uparrow$) & MSE ($\downarrow$) & F1 ($\uparrow$) & MSE ($\downarrow$)   \\  \midrule
		\texttt{SBIC}       &   {61.5}   &  {0.309}  &    \textbf{{66.0} } &  \textbf{{0.086} }   \\
		\texttt{SChem101}       &   {0.0}   &  {0.905}  &    \textbf{{52.3} } &  \textbf{{0.056} }   \\
		\texttt{Dilemmas}       &   {0.0}   &  {0.330}  &    \textbf{{34.2} } &  \textbf{{0.165} }   \\
		\texttt{DynaSent}       &   \textbf{{74.9}}   &  {0.361}  &    {11.8}  &  \textbf{{0.114}}   \\
		\texttt{Politeness}       &   {55.9}   &  {0.490}  &    \textbf{{56.8} }  & \textbf{0.110} \\
                \bottomrule
	\end{tabular}
    % \end{adjustbox}
    \end{center}
    \label{table:disagreement_prediction}
\end{table}

\begin{table*}[t]
    \caption{Evaluation results of vanilla (RoBERTa) classifiers \textbf{with text and demographics inputs} on the SBIC and Social Chemistry datasets. Note that only these two datasets among the five include the demographics of the annotators. }
    \vspace{-4mm}
	\begin{center}
	\begin{adjustbox}{width=1\linewidth}
	\begin{tabular}{@{}c ll|cccc@{}}	
        \multicolumn{3}{c}{} &  \multicolumn{2}{c}{\textbf{Group of demographics}} & \multicolumn{2}{c}{\textbf{Personal level demographics}}\\
        \toprule
		\textbf{Datasets} & \textbf{Input Setup} & \textbf{Label Type} & F1 ($\uparrow$)  & MSE ($\downarrow$)       & F1 ($\uparrow$)  & MSE ($\downarrow$)     \\  \midrule
  \multirow{2}{*}{\texttt{SBIC}}
%  &  Demographics(sentence) ; Text & Binary   &   {61.4}    &  {0.299}     &  {73.6} &  {0.202}      \\
%		 &  Demographics(templated) ; Text & Binary      &   {61.7}    &  {0.299}     &  {76.5}   &  {0.181}      \\
		 &  Demographics(sentence) ; Text & Continuous     &   {65.4}    &  {0.086}     &  \textbf{{85.6}  } &  \textbf{{0.033}  }    \\
		 &  Demographics(templated) ; Text & Continuous       &   {64.6}    &  {0.087}  &  \textbf{{85.6}  } &  \textbf{{0.033}  }  \\\midrule
		\multirow{2}{*}{\texttt{SChem101}}  
       %     &  Demographics(sentence) ; Text & Binary   &   {95.0}    &  {0.100}     &  {94.0}   &  {0.106}      \\
	%	  &  Demographics(templated) ; Text & Binary      &   {0.0}    &  {0.904}   &    {87.0}   &  {0.211}      \\
		  &  Demographics(sentence) ; Text & Continuous     &   {14.6}    &  { 0.065}     &  \textbf{{48.8}  } &  \textbf{{0.009}  }    \\
		 &  Demographics(templated) ; Text & Continuous       &   {16.9}    &  {0.073}     &  \textbf{{50.7}  } &  \textbf{{0.009}  }  \\
                \bottomrule
	\end{tabular}
    \end{adjustbox}
    \end{center}
    \label{table:demographic_disagreement_pred}
\end{table*}

\begin{table*}[!]
    \small
    \centering
    \caption{Examples of four types of predictions: (1) the predictions are the same as the collected disagreement, (2) both predictions are the same but differ from the collected disagreement. When the prediction given only text differ from the given text with demographics, (3) the former is close to collected disagreement, and (4) the latter is close to collected disagreement. }    
    \begin{tabular}{@{}c@{\hskip 1mm} p{9.5cm}@{}c@{} c @{} c@{}}
        \toprule
       \textbf{Type} &\textbf{Input} & \makecell{\textbf{Predicted} \\\textbf{Disagreement}} & \makecell{\textbf{Predicted Disagreement} \\\textbf{with Demographics}} & \makecell{\textbf{Collected}\\ \textbf{Disagreement}}  \\
    \toprule
      (1) & age: 22, politics: liberal, race: white, gender: \colorbox{NavyBlue!14}{woman}. [SEP] what is the \colorbox{NavyBlue!7}{most} \colorbox{NavyBlue!21}{common} name for a \colorbox{NavyBlue!84}{black} \colorbox{BurntOrange!42}{dude} \colorbox{NavyBlue!77}{king} kong. & 0.003 & 0.015 & 0\\
    \midrule
      (2) & age: 25, politics: liberal, race: asian, gender: woman. [SEP] swear to god \colorbox{NavyBlue!48}{they} \colorbox{BurntOrange!27}{are} the same \colorbox{BurntOrange!54}{people} that \colorbox{BurntOrange!57}{objectify} women \colorbox{BurntOrange!27}{and} reify rape \colorbox{BurntOrange!84}{culturemisogyny}. More importantly y do they feel so threatened. & 1.034 & 0.971 & 1\\
    \midrule
      (3) & age: \colorbox{NavyBlue!40}{18}, politics: other, race: asian, gender: \colorbox{NavyBlue!60}{man}.  [SEP] \colorbox{BurntOrange!40}{two} strong \colorbox{BurntOrange!80}{men} needed to \colorbox{NavyBlue!40}{lift} some \colorbox{NavyBlue!80}{furniture}. & 0.223 & 0.096 & 0 \\
    \midrule
      (4) & age: 29, politics: liberal, race: white, gender: woman. [SEP] 13 reasons why is making \colorbox{NavyBlue!21}{me} \colorbox{NavyBlue!12}{wish} i went \colorbox{NavyBlue!9}{to} a \colorbox{NavyBlue!27}{real} \colorbox{BurntOrange!78}{fucking} \colorbox{NavyBlue!84}{highschool}.  & 0.332 & 0.01 & 0.5 \\
    \bottomrule
    \end{tabular}
    \label{tab:pred_examples}
\end{table*}

\subsection{Experimental Details}\label{sec:4.2}

% \textbf{Training details.} 
All the experiments are conducted by fine-tuning RoBERTa-base \citep{liu2019roberta} using Adam optimizer \citep{kingma2015adam} with a fixed learning rate 1e-5 and the default hyperparameters of Adam.
For the text classification tasks, the model is fine-tuned with batch size 8 for 15 epochs. 

To the best of our knowledge, we couldn't find any existing disagreement predictors to be used as baselines. 
As a result, we compare our predictors with different input types and disagreement labeling setups.
Different versions of pre-trained language models were tested, but RoBERTa always performed better.
% \textbf{Evaluation.} 
For the evaluation of the performance of the trained disagreement predictor, we use both 1) hard score F1 and 2) soft score Mean Square Error (MSE), and compare the measurement effect of binary disagreement label and continuous disagreement rate. 
% Also, F1 and MSE evaluate the performance of language models in disagreement prediction.  

\subsection{Main Results}\label{sec:4.3}
\paragraph{Disagreement prediction only with text}
From Table \ref{table:disagreement_prediction}, we notice that continuous disagreement achieves better prediction than binary disagreement for most of the datasets. 
Among the datasets, the disagreement prediction models work the best in the SBIC dataset.  The binary label prediction are close to continuous prediction for SBIC and Politeness datasets. But SChem and Dilemmas have 0 F1 scores which only give 0 outputs. That means the binary label is not reliable for the two datasets.  

For Dynasent, the binary label has an inconsistent performance based on hard score F1 and soft score MSE. We think one potential reason is that the binary disagreement is highly unbalanced while converting a continuous prediction to categorical labels like 0, 0.33, 0.67, and 1 is easy to accidentally assign an intermediate value to a wrong group. Therefore, even though we used both F1 and MSE metrics, they are used to have a parallel comparison between the binary label and continuous label setup. Among the binary classification, we consider F1 as the metric of model goodness, on the opposite, we use MSE to evaluate the regression fitness.  

\paragraph{Disagreement prediction with text and demographic information}
Further, by comparing different experiment setups for disagreement with demographic information in Table \ref{table:demographic_disagreement_pred}, we focus on the different effects of a group of demographics or the personal level of demographics. The results show that personal-level demographics improve the disagreement prediction more than group-level demographics. One potential reason is that the annotator's level of demographics may imitate the annotation process that each annotator labels the text without knowing each other. And also because concatenating personal level demographics can be considered as oversampling that group-level setup can not.

\begin{figure*}[t]
\begin{center}
    {
%    \subfigure[SBIC Dataset]
%        {
%        \includegraphics[width=0.45\textwidth]{Figures/sbic_v3.pdf} 
%        \label{fig:SBIC_simulation}
%        }
%    \hfill
%    \subfigure[Schem101 Dataset]
%        {
%        \includegraphics[width=0.45\textwidth]{Figures/schem_v3.pdf}
%        \label{fig:SChem_simulation}
%        } 
\includegraphics[width=1\textwidth,trim={0.7cm 0.6cm 0.7cm 0.7cm},clip]{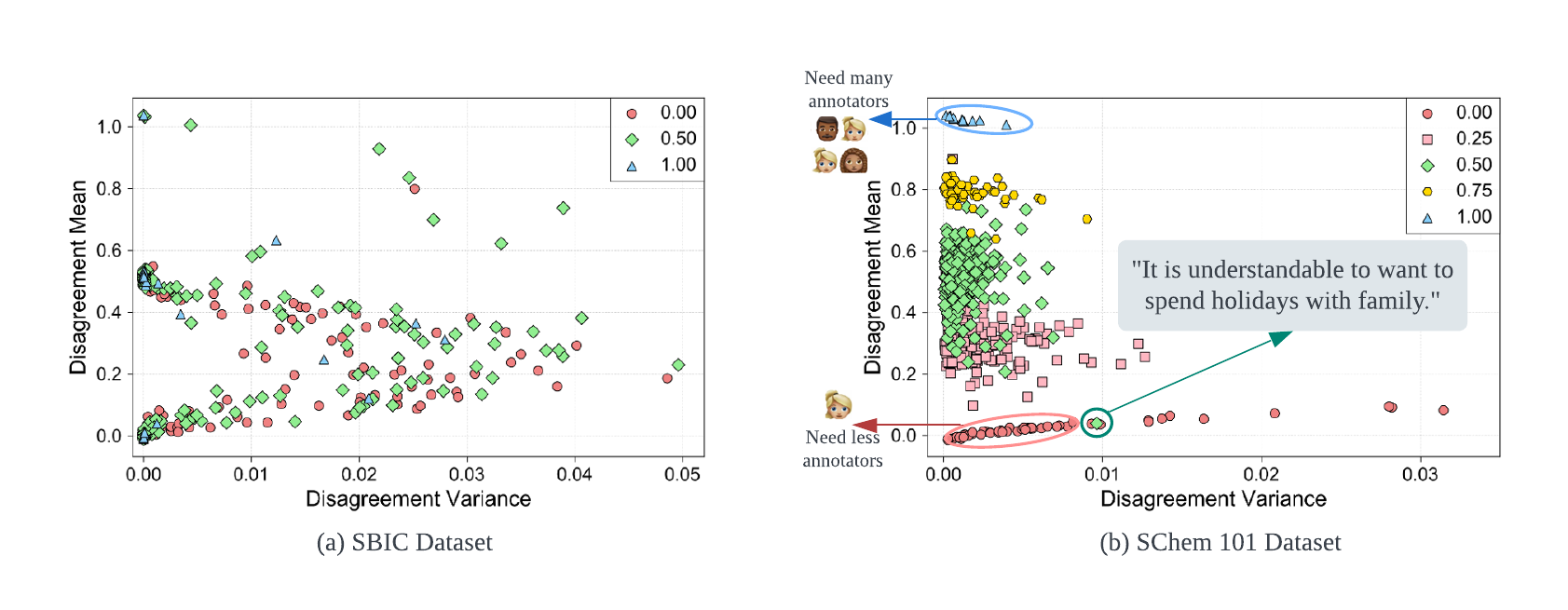}
    }
\end{center}
\vspace{-3mm}
\caption{Disagreement prediction with simulated demographic information on (a) SBIC and (b) Schem101 datasets, respectively. Different shapes and colors indicate the different disagreement labels as denoted in the legend. Best viewed in color.}
\label{fig:simul}
\end{figure*}

\paragraph{Qualitative Results Analysis}
Lastly, we categorize prediction into four types and provide an example per each in Table \ref{tab:pred_examples}.
%The first example had a consistent disagreement of 0. Second example were predicted disagreement of 1 from both setups, but the collected disagreement is 0. The trained language model might consider the words used are similar to controversial text. When the prediction given text with demographics differs from the text-only setup, the instance is sensitive to demographics. For the third example, the model given only text predicted 0.223, but the model given text with demographics predicted 0.096, close to the collected disagreement of 0. 
Using Local Interpretable Model-Agnostic Explanations (LIME) \cite{lime}, we found demographics have been used in prediction with the text. Blue text is important feature to predict agreement while orange text is used to predict disagreement. 
\subsection{Simulation of Everyone's Voices with Artificial Demographics}\label{sec:4.5}
One remaining question is how to reflect everyone's diverse opinions on such subjective and socially sensitive annotation tasks. To explore this aspect, we run additional experiments with the simulated demographics introduced in Section \ref{sec:3.3}. Namely, we simulate a different combination of all possible \textit{artificial} demographic groups, rather than using the real annotators' demographics used in model training (Section \ref{sec:4.3}). Then, the disagreement of the simulated demographic information and the text is predicted using the fine-tuned disagreement predictor introduced in Section \ref{sec:3.2}.

% We designed our study motivated by the Intersectionality theory: different people's occasions are shaped by the intersection of multiple forms of discrimination \cite{crenshaw1990mapping}. 
Our study is motivated by the Intersectionality theory \cite{crenshaw1990mapping}, assuming that people's perspectives are shaped by the intersection of all available demographic categories. We set four gender types, seven ethnicity types, and five age ranges (see \ref{appendix:Simulation Setup} for details), and thus we have 140 (4$\times$7$\times$5) artificial annotators' unique demographic characteristics. 
% Specifically, we set gender with \textit{four} possible options: woman, man, transgender, non-binary; ethnicity with \textit{seven} options: white, black or African American, American Indian or Alaska Native, Asian, Native Hawaiian or other pacific islanders, Hispanic, or some other race. Also, we set the age with \textit{five} ranges: 18 to 29, 30 to 39, 40 - 49, 50-59, and 60 to elder. 
Since we only trained our disagreement predictor with demographic information on SBIC and SChem101 datasets, the simulation experiments are also applied to these two datasets. We randomly sampled 600s text instances in each dataset and concatenated them with 140 artificial annotators' demographic information in the colon template to predict continuous disagreement. 

%Figure \ref{fig:simul} showed that highly disagreed text has a smaller variance among the 140 equivalent representative artificial annotators. Also, a few texts from the SBIC changed their continuous disagreement significantly. For instance, one text has a disagreement label as 0 from the original annotation, but 1.03 as the mean disagreement of the text among 140 artificial annotators, e.g., the text is : ``i have news... the terrorist barack hussein obama has been hunted down and killed... girlish giggle oops i meant osama bin laden.'' 
%On the other hand, one text from the SChem101 dataset has an original 0.5 disagreement label but ends up with a 0.04 disagreement prediction among 140 artificial annotators, e.g., the text is ``It is understandable to want to spend holidays with family.'' 
%These two examples illustrate that an extremely increased or decreased disagreement score from the simulation experiment can indicate these texts need further real annotation to determine a more acceptable answer. 
To visualize the simulation result of 140 artificial annotators per text, we made a scatter plot based on the mean and variance of 140 disagreement prediction as shown in Figure \ref{fig:simul}. The color and shape denoted at the legend shows the text's disagreement label in the original dataset. The higher points in the plot means higher predicted disagreement rate. The more rightward point implies a greater variance in the disagreement prediction among the 140 artificial annotators.
The difficulty of disagreement prediction is related to the dataset's topic, quality etc. SBIC is collected from social media data while SChem is created by crowdsourcing, which might explain why the clusters are more clear in the Figure \ref{fig:simul}(b) than in the Figure \ref{fig:simul}(a). From \ref{fig:simul}(b), most text are predicted into corresponding disagreement clusters. But some outliers are predicted to be more controversial or agreeable. For example, the circled outlier has an original 0.5 disagreement label but ends up with a 0.04 disagreement prediction among 140 artificial annotators. The text is "It is understandable to want to spend holidays with family." Those outliers in the simulation experiment show the disagreement rate would change if the annotator change. Other than the outliers, the disagreement clusters shows they are less influenced by annotator change. With this simulation, we can distinguish disagreements caused by the natural controversy of the text or by the biased distribution of the assigned annotators.

% this annotators' disagreement simulation could be used to identify the text that receives substantial divergent disagreement labels between the small size of human annotators and the large set of artificial annotators, which is a signal for requiring further context explanation or human annotation.  

\section{Discussion and Future Work}
% Based on experiments across five datasets, we find our disagreement predictor can perform well in multiple subjective tasks, such as offensiveness detection and sentiment analysis. 
% With annotators' demographic information, the disagreement predictor can perform even better. 
% Using our disagreement prediction framework, we can make the annotation process more efficient by flexibly allocating more or fewer number annotators depending on the level of disagreement.

We could think of potential applications in NLP data annotation pipeline using our disagreement prediction model:

\textbf{Annotator number estimation.}
We could potentially use the predicted disagreement score in order to decide the appropriate number of annotators in a cost-efficient manner, e.g., we may not need three or five annotators for the text being predicted zero disagreements.
For instance, we may need one or two annotators if a text is predicted to have lower disagreement scores.
Other than that, we can assign five or even more annotators to those texts being predicted as highly disagreeable.

\textbf{Annotator group assignment.}
Additionally, we suggest considering the annotation disagreement as a critical factor in finding the optimal group of annotator pools. This can be used as a novel annotator assignment supporting system for the data annotation pipeline. In the current annotator recruiting process, there is usually some uncontrollable randomness from annotators, either from skewed representatives or individual variations. We present a low-cost approach to simulate as diverse as possible artificial annotation pools to identify the controversial samples that maximize the disagreement. 
Thus, we avoid ignoring human bias and listening to opinions from a more diverse group of people to avoid polarized analysis.  
% Although this is only applied in public datasets to evaluate collected disagreement, we plan to implement disagreement into the subjective tasks and compare the final results in future work. 
We hope our study can evoke others' attention in designing a more fair and representative annotation pipeline.  

\textbf{Potential risk of using demographic information.}
Last but not least, though our research shows that annotators' demographics help disagreement prediction, we should be careful about collecting private and personal information.   
Also, we admit that NLP or AI systems trained on demographic information might make another bias toward certain demographic groups.

\section{Conclusion}
Overall, we propose a disagreement prediction framework that measures annotators' disagreement in subjective tasks, predicts disagreement with/without demographic information and simulates 140 artificial annotators to build a relatively fair annotation pool. 
Our results show that the annotators' disagreement could be fairly predictable from the text and even better performs when we know the demographic information of the annotators.
With our disagreement predictor, we believe we could shed light on various applications of data annotation in a more effective and inclusive manner.

% Use \bibliography{yourbibfile} instead or the References section will not appear in your paper
\section*{Acknowledgments}
We thank Dr. Maxwell Forbes for sharing the demographic information data for Social Chemistry 101 dataset.
We also thank the anonymous reviewers and Minnesota NLP members for their insightful comments and suggestions.
\bibliography{aaai23}

\clearpage
\appendix
\section{Appendices}

\subsection{Simulation Setup of Artificial Annotators}\label{appendix:Simulation Setup}
We set gender with \textit{four} possible options: woman, man, transgender, non-binary; ethnicity with \textit{seven} options: white, black or African American, American Indian or Alaska Native, Asian, Native Hawaiian or other pacific islanders, Hispanic, or some other race. Also, we set the age with \textit{five} ranges: 18 to 29, 30 to 39, 40 - 49, 50-59, and 60 to elder.

%\subsection{Supplementary of Table \ref{table:demographic_disagreement_pred}}
%Table \ref{table:demographic_disagreement_pred_binary} shows the evaluation results of RoBERTa with text and demographics inputs, predicting binary labels on the SBIC and Social Chemistry datasets.

\subsection{Annotators Distributions}
Our analysis finds that the annotators' pool in the SBIC dataset was relatively gender-balanced and age-balanced (55\% women, 42\% men, 1\% non-binary; 36±10 years old), but racially skewed (82\% White, 4\% Asian, 4\% Hispanic, 4\% Black). And it was also politically skewed (63\% liberal, 20\% conservative). Overall, workers agreed on a post being offensive at a rate of 76\%. Later, \citeauthor{Sap2020SocialBF} showed that annotator identity and beliefs are highly related to their toxicity ratings in their annotators with attitudes paper \cite{sap-etal-2022-annotators}. 
Similar to the demographic distribution in the SBIC dataset, the crowd worker pool in SChem101 is also gender-balanced and race-skewed: 55\% were women and 45\% men. 89\% of workers identified as white, 7\% as Black. 39\% were in the 30-39 age range, 27\% in the 21-29, and 19\% in the 40-49 age range. Regarding education, 44\% had a bachelor's degree, and 36\% had some college experience or an associate's degree. However, even though some people consider one rule as a common belief, other people may think no one believes it.

%\subsection{Disagreement Distributions}
%\begin{figure}[h]
%\centering
%\includegraphics[width=0.5\textwidth,trim={0.6 0 0.7cm 0.7cm},clip]{Figures/disagreement_distribution.pdf}
%\caption{Disagreement distributions for five datasets}
%\label{fig:disagreement_distribution}
%\end{figure}
%The Figure \ref{fig:disagreement_distribution} shows the distributions of disagreement scores among five datasets. For dynasent dataset, since the majority of the dataset has disagreement between 0.3 to 0.6. The prediction concentrate around 0.4 to 0.5. The comparison among multiple datasets reflects that the subject topics influence the crowd annotators' disagreement. For example, most texts regarding offensiveness had consensus opinions from the annotators, while most annotators disagreed regarding sentiment.

\subsection{Group v.s. Personal Demographics Setup}
Table \ref{tab:Group v.s. Person } shows one example of text with individual annotators' demographics or with the group of annotators' demographics.
\begin{table}[t]
    
    \centering
    \caption{Setups between group and personal demographics}   
    \begin{tabular}{c c}
        \toprule
       Input & Disagreement Label  \\
    \toprule
      \makecell{age: 22, politics: liberal,\\race: white, gender: woman.\\ what is the most common \\ name for a black dude king kong.} & 0\\
     \cdashline{1-2}
     \makecell{age: 34, politics: liberal,\\ race: white, gender: woman.\\what is the most common\\name for a black dude king kong} & 0 \\
     \cdashline{1-2} 
     \makecell{age: 57, politics: mod-liberal,\\ race: white, gender: woman.\\what is the most common \\name for a black dude king kong} & 0 \\
     \midrule
     \makecell{age: 22, politics: liberal,\\race: white, gender: woman.\\age: 34, politics: liberal,\\ race: white, gender: woman.\\age: 57, politics: mod-liberal,\\ race: white, gender: woman.\\what is the most common \\name for a black dude king kong} & 0 \\
    \bottomrule
    \end{tabular}
    \label{tab:Group v.s. Person }
\end{table}

\subsection{Disagreement Prediction Given Only Or Partial Demographics}
\begin{table}[h!]
    \caption{Evaluation results of vanilla (RoBERTa) classifiers with only demographics 
 or text and partial demographics.}
    \vspace{-4mm}
	\begin{center}
	\begin{adjustbox}{width=1\linewidth}
	\begin{tabular}{@{}r l|cc@{}}	
        \multicolumn{2}{c}{}  & \multicolumn{2}{c}{\textbf{Personal level demographics}}\\
        \toprule
		\textbf{Datasets} & \textbf{Input Setup} & F1 ($\uparrow$)  & MSE ($\downarrow$)     \\  \midrule
      \multirow{1}{*}{\texttt{SBIC}}
      &  Only demographics & {32.29} &  {0.172 } \\
   %   &  Age; text & { } &  { } \\
   %   &  Politics; text & { } &  { } \\
   %   &  Race; text & { } &  { } \\
   %   &  Gender; text & { } &  { } \\
      \midrule
      \multirow{4}{*}{\texttt{SChem}}
      &  Only demographics   &   {14.0} &  {0.134} \\
      & Age; text &  {49.39} &  {0.008} \\
      & Education; text & {47.04} & {0.009} \\
      &  Race; text &  {47.04} &  {0.009} \\
      &  Gender; text & {47.04} &  {0.008}\\    
                \bottomrule
	\end{tabular}
    \end{adjustbox}
    \end{center}
    \label{table:partial_only_demographic_disagreement_pred}
\end{table}

To further evaluate how annotators' demographics influence disagreement prediction, we also tested the inputs of only demographics, which performed much worse than the inputs including text. Notably, this experimental input setup might mislead, assuming people from certain social groups always have the kind of opinion regardless of the text context. 

Based on our above study, we controlled the demographics in the templated format of individual annotators and the label in the continuous format, which is the optimal setup. And we evaluated the input of text with partial demographic information as shown in Table \ref{table:partial_only_demographic_disagreement_pred}. It shows that the predictions are given input of text with a single demographic factor, or only demographics perform worse than predicting with text and intersectional demographic information. We also tried using random forests given only demographic features to predict annotation disagreement. The age feature was the most important.
\subsection{Results of Other Language Models on Disagreement Prediction}
We only reported Roberta in our main paper, which showed the best performance. But we have also conducted experiments with other language models like BERT\cite{devlin2018bert}, XLNet\cite{yang2019xlnet}, and AlBERTa\cite{lan2019albert}. Table \ref{table:bert_disagreement_prediction} shows the other language models' prediction results on SChem as an example. 
\begin{table}[!t]
    \caption{Evaluation results of different classifiers with text input on SChem Dataset with disagreement.}
    \vspace{-4mm}
	\begin{center}
	\begin{adjustbox}{width=0.9\linewidth}
	\begin{tabular}{c cc}	
        \multicolumn{1}{c}{} &   \multicolumn{2}{c}{Text with Continuous Label}\\
        \toprule
		{Model} & F1 ($\uparrow$) & MSE ($\downarrow$)   \\  \midrule
		\texttt{BERT} & {25.18} & {0.061}\\
		\texttt{XLNet} & {20.22} &  {0.068}\\ 
        \texttt{AlBERTa} & {20.81} & {0.066} \\
        \bottomrule
	\end{tabular}
    \end{adjustbox}
    \end{center}
    \label{table:bert_disagreement_prediction}
\end{table}

\end{document}